\pdfoutput=1

\documentclass[11pt]{article}

\usepackage{emnlp2021}

\usepackage{times}
\usepackage{latexsym}
\usepackage[tikz]{bclogo}

\usepackage[T1]{fontenc}

\usepackage[utf8]{inputenc}

\usepackage{microtype}

\usepackage{times}
\usepackage{graphicx}
\usepackage{caption}
\usepackage{subcaption}
\usepackage{wrapfig}
\usepackage{dsfont}
\usepackage{latexsym}
\usepackage[linesnumbered,algoruled,boxed,noend]{algorithm2e}
\usepackage{amssymb}
\usepackage{amsmath}
\usepackage{url} 
\SetKwInOut{Parameter}{parameter}
\usepackage{enumitem}
\usepackage{mathtools}
\usepackage{cleveref}
\usepackage{multirow}
\usepackage{hhline}
\usepackage[export]{adjustbox}
\usepackage{booktabs,array}
\usepackage{amsmath, bm}
\usepackage{color, colortbl}
\usepackage{xcolor}
\usepackage{resizegather}
\usepackage{fixltx2e}

\SetKwInput{KwInput}{Input}
\SetKwInput{KwOutput}{Output}
\newcolumntype{P}[1]{>{\centering\arraybackslash}p{#1}}
\usepackage{booktabs,makecell,tabularx} 
\setitemize{noitemsep,topsep=0pt,parsep=0pt,partopsep=0pt}
\let\oldnl\nl
\definecolor{Gray}{gray}{0.9}
\definecolor{LightCyan}{rgb}{0.88,1,1}
\newcommand{\nonl}{\renewcommand{\nl}{\let\nl\oldnl}}

\usepackage{pifont}

\usepackage{epigraph}

\usepackage{etoolbox}
\patchcmd{\epigraph}{\@epitext{#1}}{\itshape\@epitext{#1}}{}{}

\newlength{\Width}%
\newlength{\DepthReference}
\newlength{\HeightReference}
\newcommand{\MyColorBox}[2][red]%
{%
    \settowidth{\Width}{#2}%
    \colorbox{#1}%
    {%
        \raisebox{-\DepthReference}%
        {%
                \parbox[b][\HeightReference+\DepthReference][c]{\Width}{\centering#2}%
        }%
    }%
}
\newcommand\rockemoji{\raisebox{-5pt}{\includegraphics[width=1.4em]{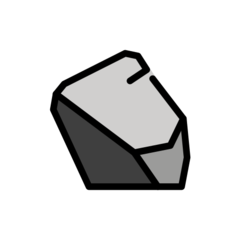}}}

\newcommand{\sectionref}[1]{\S\ref{#1}}

\definecolor{msftBlue}{RGB}{0,164,239}
\definecolor{msftGreen}{RGB}{127,186,0}
\definecolor{msftYello}{RGB}{255,185,0}
\definecolor{msftBlack}{RGB}{0,0,0}
\newcommand{\finding}[1]{
	\begin{bclogo}[couleur= msftBlue!15,  arrondi=0, logo=\bclampe, marge=2,   couleurBord=msftBlue!10,  sousTitre ={\em \textit{\textbf{#1}}}]{} 
	\end{bclogo}
	\vspace{-1.4em}
}

\title{\vspace*{-0.5in}
{{\small \hfill \textit{in Proc. of EMNLP 2021 (short)}}\\
\vspace*{.25in}}\rockemoji{}RockNER: A Simple Method to Create Adversarial Examples \\for Evaluating the Robustness of Named Entity Recognition Models}

\author{
Bill Yuchen Lin \quad Wenyang Gao \quad Jun Yan \quad Ryan Moreno  \quad Xiang Ren\\
\texttt{\{yuchen.lin,wenyangg,yanjun,morenor,xiangren\}@usc.edu}\\
Department of Computer Science and Information Sciences Institute,  \\ University of Southern California\\
}

\begin{document}

\maketitle

\begin{abstract}
To audit the robustness of named entity recognition (NER) models, we propose RockNER, a simple yet effective method to create natural adversarial examples. 
Specifically, at the entity level, we replace target entities with other entities of the same semantic class in Wikidata;
at the context level, we use pre-trained language models (e.g., BERT) to generate word substitutions. 
Together, the two levels of attack produce natural adversarial examples that result in a shifted distribution from the training data on which our target models have been trained.
We apply the proposed method to the OntoNotes dataset and create a new benchmark named \texttt{OntoRock} for evaluating the robustness of existing NER models via a systematic evaluation protocol. 
Our experiments and analysis reveal that even the best model has a significant performance drop, and these models seem to memorize in-domain entity patterns instead of reasoning from the context.
Our work also studies the effects of a few simple data augmentation methods to improve the robustness of NER models. \footnote{Our code and data are publicly available at the project website: \url{https://inklab.usc.edu/rockner}.}

\end{abstract}
\section{Introduction}
\label{sec:intro}

Recent named entity recognition (NER) models have achieved great performance on many conventional benchmarks such as CoNLL2003~\cite{conll} and OntoNotes 5.0~\cite{weischedel2013ontonotes}.
However, it is not clear whether they are reliable in realistic applications in which entities and/or context words can be out of the distribution of the training data.
It is thus important to audit the robustness of NER systems via natural adversarial attacks.
Most existing methods for generating adversarial attacks in  NLP focus on sentence classification 
\citep{jin2020bert, Li2020BERTATTACKAA, minervini2018adversarially} and question answering~\citep{jia2017adversarial, ribeiro-etal-2018-semantically, gan2019improving},
and these methods lack special designs reflecting the underlying compositions of the NER examples --- i.e., \textit{entity} structures and their \textit{context} words. 
In this paper, we focus on creating general natural adversarial examples (i.e., real-world entities and human-readable context) for evaluating the robustness of NER models.

\begin{figure}[t] 
	\centering 
	\includegraphics[width=0.95\linewidth]{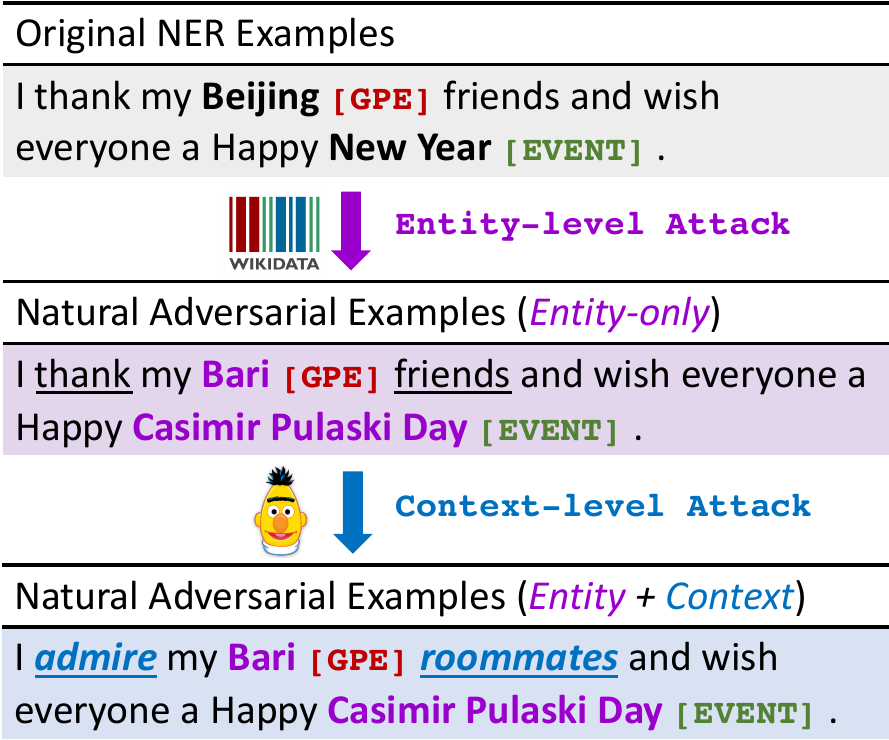}
	\caption{Illustration of RockNER attacking pipeline. }
	\label{fig:example} 
\end{figure}

As shown in Figure~\ref{fig:example}, given a NER example,
our method first generates entity-level attacks by replacing the original entities with entities from Wikidata and then uses a pre-trained masked language model like BERT~\cite{Devlin2019} to generate context-level attacks.
We choose the OntoNotes dataset~\cite{weischedel2013ontonotes}\footnote{The proposed attacking method is also applicable to other general NER datasets.} to showcase \textsc{RockNER} because of the dataset's high annotation quality and wide coverage of entity types. Thus, we create a novel benchmark, named \texttt{OntoRock}, for evaluating the robustness of a wide range of modern NER models.  
%

We analyze the robustness of popular existing NER models on the \texttt{OntoRock} benchmark in order to answer three research questions as follows: 
(Q1) How robust are current NER models? 
(Q2) Where are the NER models brittle? 
(Q3) Can we improve the robustness of NER models via data augmentation?
Our experiments and analysis provide these main findings:
1) even the best model is still brittle to our natural adversarial examples, resulting in a significant performance drop (92.4\% $\rightarrow$ 58.5\% in F1); 
2) current NER models tend to memorize entity patterns instead of reasoning based on the context; also, there are specific patterns for entity typing mistakes;
3) simple data augmentation methods can indeed help us improve the robustness to some extent.
We believe the proposed RockNER method, the OntoRock benchmark, and our analysis will benefit future research to improve the robustness of NER models.




\section{Natural Adversarial Attacks for NER} 
We present RockNER, a simple yet effective method to generate high-quality natural adversarial examples for evaluating the robustness of NER models by perturbing both the entities and contexts of original examples.
We apply the method to the development set and test set of OntoNotes to create the \textbf{\texttt{OntoRock}} benchmark.


\subsection{Entity-Level Attacks}
\label{subsec:entity_attacks}
\begin{figure}[t]
	\centering 
	\includegraphics[width=0.90\linewidth]{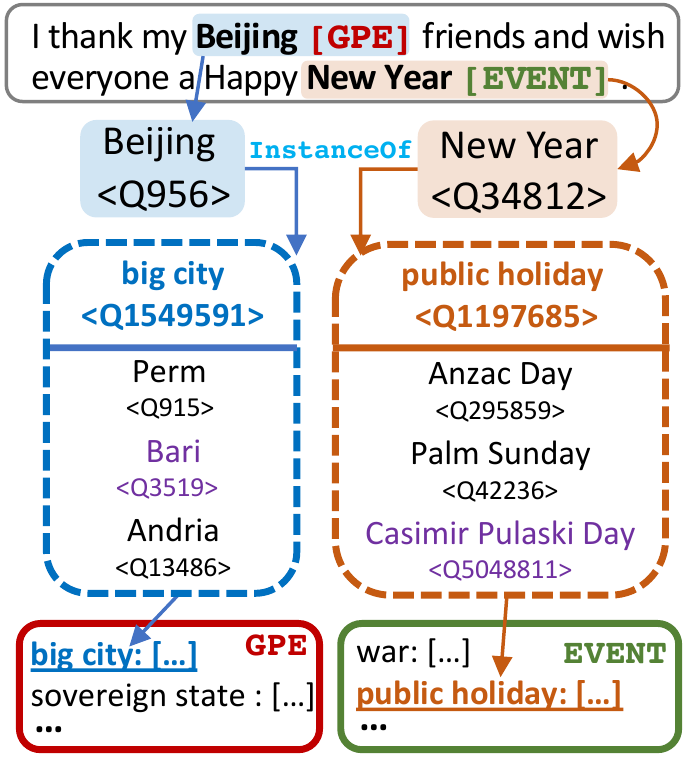}
	\vspace{-0.4em}
	\caption{Building adversarial entity dictionary. \vspace{-0.5em}}
	\label{fig:entityattack} 
\end{figure}

To generate relevant entities for modifying existing NER data,  we collect a dictionary of natural adversarial entities of different fine-grained classes via Wikidata.
As shown in Figure~\ref{fig:entityattack}, 
our three-stage pipeline is introduced as follows:
\begin{itemize}
    \item \textbf{(1)} \textit{Entity Linking}:
We first use \textit{BLINK}~\cite{wu2019zero} to link each entity in the original examples from its surface form to a canonical entry in Wikidata with a unique identifier (QID), e.g., ``Beijing'' $\rightarrow Q956$. 
    \item {\textbf{(2) }\textit{Fine-grained Classification}:}
Then, we execute a query to get its fine-grained class via the \texttt{InstanceOf} relation (P31), e.g., $Q956\xrightarrow[]{P31}Q1549591$ (``big city''). 
    \item \textbf{(3)} \textit{Dictionary Expansion:}
Finally, we retrieve additional Wikidata entities within each individual entity class. Given a particular entity such as ``Beijing'', we collect additional \textit{out-of-distribution} entities, such as ``Bari''. 
They are both big cities (\textsc{Gpe} type), while the latter one is \textit{much less correlated} with the context in the training data\footnote{We assure this by applying tested NER systems (\S\ref{sec:exp}) on sentences where each sentence is the an individual entity name, and only keep the ones the target models predict incorrectly.}.
\end{itemize}



\quad 
To ensure the quality,
we manually curate the fine-grained entity classes and remove entities linked incorrectly.
We use a different approach for collecting \textsc{Person} attack entities because adversarial names can be more efficiently created as random combinations of first names, middle names, and last names, which are collected from the Wikidata person-name list\footnote{{We used the method described in \url{https://github.com/EdJoPaTo/wikidata-person-names}}}.







To create an evaluation benchmark based on existing datasets (e.g., OntoNotes),  we iterate over every original entity and replace it with a randomly sampled adversarial entity from our dictionary sharing the \textit{same fine-grained class}.
We argue that the resulting attacks are both \textbf{\textit{natural}} --- i.e., containing real, valid entities, and \textbf{\textit{adversarial}} --- i.e., the entities are of the same class as the original entities while being out-of-distribution from the training data.
Specifically,
\texttt{OntoRock} has a much larger vocabulary of entity words than OntoNotes, and these words are rarely seen in the training set (Table~\ref{tab:outofdist} in Appendix \sectionref{app:dataset_stats}).
For example, for \textsc{Gpe} and \textsc{Product}, \texttt{OntoRock} has $\sim$3x the number of unique entity words as OntoNotes has (\textsc{Gpe}: $461$ vs. $1202$, \textsc{Product}: $54$ vs. $158$). The ratio of seen entity words is also much lower
 (\textsc{Gpe}: $75.92\%$ vs. $17.30\%$, \textsc{Product}: $44.44\%$ vs. $7.59\%$).



\label{sec:method}

\subsection{Context-level Attacks}
\label{subsec:context_attacks}
To investigate the robustness of NER models against changes to the context,
we also create natural attacks on the context words.
Our intuition is to replace context words with words that are semantically related and syntactically valid but out of the distribution of the training data.
To this end, 
we perturb the original context by sampling adversarial tokens via a masked-language model such as BERT.
Specifically, 
for each sentence, we choose semantic-rich words --- nouns, verbs, adjectives, and adverbs --- as the target tokens to replace. 
Then, we generate masked sentences with random numbers (at most 3) of \textsc{[mask]} tokens. 
These masked sentences are then fed into BERT, which decodes the masked positions one by one from left to right.
We use the predicted tokens ranking between 100$\sim$200, such that the words create more challenging context yet the sentence is still syntactically valid.
As there are multiple sampled sentences, we take the one which is the least correlated with the training data.
Specifically, we test all candidate sentences on the trained BLSTM-CRF model (which performs the worst among the target NER models) and we select the sentences that cause a performance drop.



\begin{table}[t!]
	\centering
	\scalebox{0.73
	}{
		\begin{tabular}{@{}r|c|c|c|c}
\toprule
& {\MyColorBox[gray!20]{\textbf{None}}} & {\MyColorBox[red!10]{\textbf{E}}} & {\MyColorBox[cyan!20]{\textbf{C}}} & {\MyColorBox[green!20]{\textbf{E+C}}}\\ 
\midrule

BLSTM-CRF & 84.6 & 40.5 \textsubscript{($\downarrow$ 52\%)} & 77.3 \textsubscript{($\downarrow$ 9\%)} & 32.4 \textsubscript{($\downarrow$ 62\%)}\\
\midrule
SpaCy & 87.3 & 43.9 \textsubscript{($\downarrow$ 50\%)} & 81.8 \textsubscript{($\downarrow$ 6\%)} & 40.1 \textsubscript{($\downarrow$ 54\%)}\\ 
Stanza & 87.9 & 56.1 \textsubscript{($\downarrow$ 36\%)} & 83.0 \textsubscript{($\downarrow$ 6\%)} & 51.7 \textsubscript{($\downarrow$ 41\%)}\\ 
\midrule
BERT-CRF & 90.6 & 59.2 \textsubscript{($\downarrow$ 35\%)} & 85.8 \textsubscript{($\downarrow$ 5\%)} & 54.6 \textsubscript{($\downarrow$ 40\%)}\\ 
Flair & 90.7 & 59.6 \textsubscript{($\downarrow$ 34\%)} & 86.1 \textsubscript{($\downarrow$ 5\%)} & 55.3 \textsubscript{($\downarrow$ 39\%)}\\
\underline{\textit{\textbf{RoBERTa-CRF}}} &  {\MyColorBox[gray!20]{\textbf{92.4}}} & {\MyColorBox[red!10]{\textbf{63.4}}} \textsubscript{($\downarrow$ 31\%)} & {\MyColorBox[cyan!10]{\textbf{87.2}}} \textsubscript{($\downarrow$ 6\%)} & {\MyColorBox[green!20]{\textbf{58.5}}} \textsubscript{($\downarrow$ 37\%)}\\ \midrule
\midrule
+ \textbf{Ent. Switch.} & 91.4 & 64.7 \textsubscript{($\downarrow$ 29\%)} & 85.7 \textsubscript{($\downarrow$ 6\%)} & 59.1 \textsubscript{($\downarrow$ 35\%)}\\
+ \textbf{Rand. Mask.} &  {{\textbf{92.6}}} & {\MyColorBox[red!10]{\textbf{66.3}}} \textsubscript{($\downarrow$ 28\%)} & 86.4 \textsubscript{($\downarrow$ 7\%)} & {\MyColorBox[green!20]{\textbf{60.0}}} \textsubscript{($\downarrow$ 35\%)} \\ 
+ \textbf{Mixing Up} & 92.0 & 61.1 \textsubscript{($\downarrow$ 34\%)} & {\MyColorBox[cyan!10]{\textbf{86.9}}} \textsubscript{($\downarrow$ 6\%)} & 56.5 \textsubscript{($\downarrow$ 39\%)}\\ \bottomrule             
\end{tabular}
	} 
	\caption{ F1 scores of models trained on OntoNotes' training data and evaluated in different settings: {\MyColorBox[gray!20]{\textbf{none}}} (original test of OntoNotes) and three variants of our \texttt{OntoRock} benchmark: ({\MyColorBox[red!10]{\textbf{E}}} for entity-only attacks, {\MyColorBox[cyan!20]{\textbf{C}}} for context-only attack, and {\MyColorBox[green!20]{\textbf{E+C}}} for the full version). Relative F1 drops shown as ($\downarrow x$). \vspace{-0.5em}}
	\label{tab:results}
\end{table}

\subsection{\texttt{OntoRock} as a Robustness Benchmark}
\label{subsec:full_attacks}
We create the most challenging version of our \textsc{RockNER} attack by applying both entity-level and context-level attacks on the original development and test sets of OntoNotes, forming our \texttt{OntoRock} benchmark.
The overall statistics of \texttt{OntoRock} are shown in Table~\ref{tab:overallstat} (Appendix~\S\ref{app:dataset_stats}), alongside the
statistics of the original OntoNotes dataset.
We showcase RockNER using OntoNotes in this paper because of the dataset's high annotation quality and comprehensive entity-type coverage. However, this method of attack is also applicable to other datasets.




\section{Evaluating Robustness of NER Models}
\label{sec:exp}



In this section, we use our \texttt{OntoRock} dataset to evaluate the robustness of 
popular NER models including spaCy~\cite{spacy}, Stanza~\cite{qi2020stanza}, Flair~\cite{akbik-etal-2018-contextual}, BLSTM-CRF~\cite{Lample2016NeuralAF}, BERT-CRF~\cite{Devlin2019BERTPO}, 
and RoBERTa-CRF~\cite{Liu2019RoBERTaAR}.
Model details are described in Appendix~\S\ref{app:model_details}.
We organize our results and analysis as three main research questions and their answers.




\finding{Q1: How robust are current NER models?}

\paragraph{Main results.} We show the F1 scores on the test sets\footnote{Full results on dev and test are reported in Appendix~\S\ref{app:full_results}.} in Table~\ref{tab:results}. 
We can see that all NER models have a significant performance drop in the attacked settings  (i.e., entity attack only, context attack only, and both); there is a 35\% $\sim$ 62\% relative decrease (in the models' F1) in the fully-attacked setting as compared to their results\footnote{All models are trained on the OntoNotes' training data. } on the original test set. 
We find the performance on the original test set is positively correlated with the robustness against our attacks.
Thus, models that perform better on in-domain data tend to be also better at handling out-of-distribution examples.


\paragraph{Pre-training \& Robustness.}
BLSTM-CRF is trained solely on the training set of OntoNotes;
The NER toolkits such as spaCy and Stanza are trained on more datasets (e.g., CoNLL03);
BERT-CRF, Flair, and RoBERTa-CRF are based on pre-trained language models.
We can see that, in terms of robustness, NER models with pre-training tend to outperform models without pre-training but with more NER data access, which outperform those trained only on the OntoNotes training set.
This observation indicates that pre-training (on corpora or other NER data) leads to better robustness, and better pre-trained models (RoBERTa vs. BERT) have a lower (relative) performance drop.
Interestingly, we find that the improved robustness from pre-training mainly comes from the improvement on the entity-level attacks, possibly because of the increased exposure to entities and increased ability to reason using context (see our 1st point in Q2).

\begin{figure}[t]
    \hspace{-0.5em}
	\includegraphics[scale=0.42]{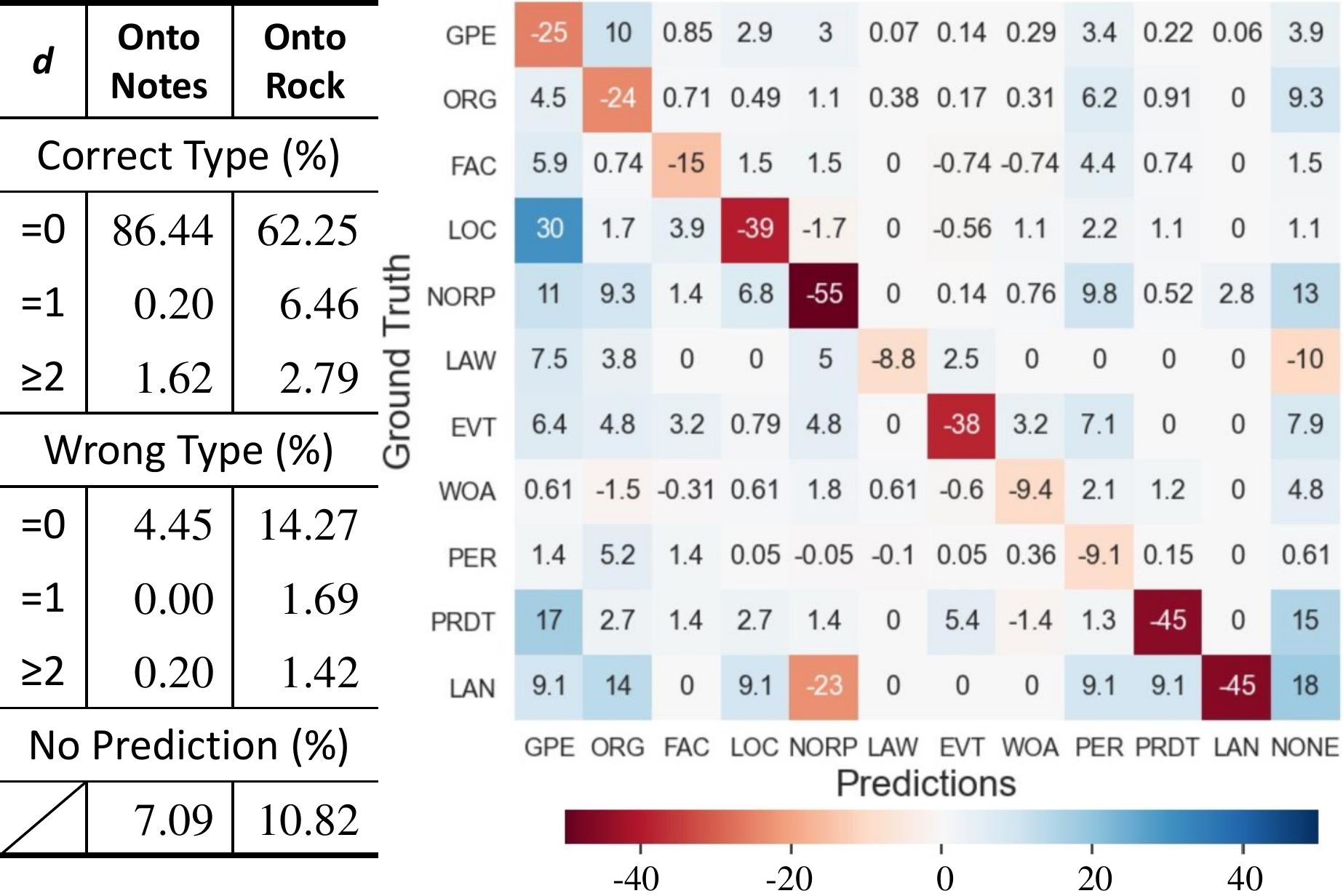}
	\caption{ Error analysis of RoBERTa-CRF. \textbf{Left}: Difference of predictions on OntoNotes and \texttt{OntoRock}. \textbf{Right}: Difference of confusion matrices. 
	\vspace{-0.5em}}
	\label{fig:error_analysis} 
\end{figure}

\finding{Q2: Where are the NER models brittle?} 

\paragraph{Memorizing or Reasoning?}
Note that our entity-level attacks aim to test the ability to use the context to infer the entities, as the novel entities themselves are out-of-distribution --- i.e., if a model can reason about the context, it should be robust against entity changes. 
In turn, the context-level attacks audit the ability to memorize entity patterns, as the context is changed, making it more challenging to infer from.
From Table~\ref{tab:results}, 
we can see that all models have a smaller performance drop in context-level attacks and a larger performance drop in entity-level attacks.
Therefore, we conclude that NER models are apt to memorize entity patterns presented in the training data and are more brittle when emerging, out-of-distribution entities exist in the inputs.
This also suggests that current NER models tend infer the type and boundary of entities without properly using the context.
To make NER models more robust,
we believe an important future direction is to develop context-based reasoning approaches, taking advantage of inductive biases such as entity triggers~\cite{Lin2020TriggerNERLW}.

\paragraph{Error Analysis.} 
To analyze the additional errors caused by our attacks,
we look at each truth entity and inspect the changes of model behaviors in this position.
We pair each original entity with its overlapped prediction and categorize it as follows:
(1) whether the predicted \textit{type} matches (Correct/Wrong);
(2) the number of \textit{different tokens} between the prediction and truth ($d$).
In Figure~\ref{fig:error_analysis} (left),
RoBERTa-CRF's predictions on OntoNotes and \texttt{OntoRock} and find that most additional error cases (86.4\% vs. 62.3\%) are caused by typing errors --- the model either predicts a wrong type (4.5\% vs. 14.3\%), or \textsc{None} (7.1\% vs. 10.8\%).
Concrete cases are shown in Figure~\ref{fig:cases} (Appendix~\S\ref{app:cases}).

We take a closer look by calculating the difference between the models' confusion matrices on the attacked and the original test data (i.e., \texttt{OntoRock}'s minus OntoNotes'), as shown in Figure~\ref{fig:error_analysis} (right).
This confusion-difference matrix reveals the model's weakness in handling novel entities, especially when making decisions between closely related categories.
For example, the biggest difference is the typing error from \textsc{Loc} to \textsc{Gpe} (increased by 30 points)\footnote{\textsc{Gpe} (Geo-Political Entity) is defined to include ``countries, cities, states'', while \textsc{Loc} (Location) is defined as ``non-GPE locations, mountain ranges, bodies of water''.} ---
indicating that the model struggles to recognize names of countries/cities/states that are not covered by the distribution of training data.

Apart from that, we find that the entity-level and context-level attacks succeed in different parts of examples.
We denote the sets of entity spans that are mistakenly predicted in entity-only attacks and context-only attacks as $S_\text{E}$ and $S_\text{C}$.
Their Jaccard similarity is only $0.04$, which shows that these two attacks target different kinds of weaknesses.

\finding{Q3: Can we improve the robustness of NER models via data augmentation?}

\begin{figure}[t]
	\centering 
	\includegraphics[width=1\linewidth]{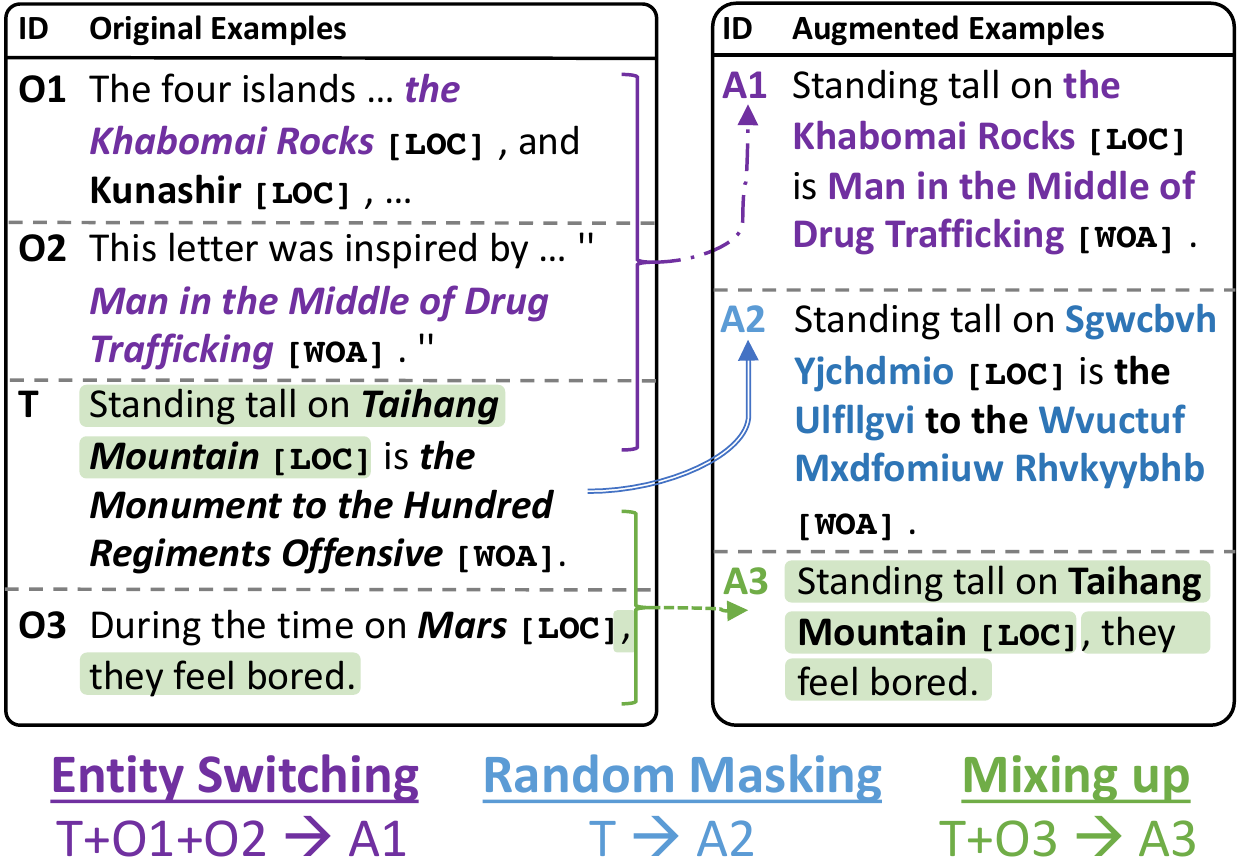}
	\caption{Three simple methods for augmenting training data to improve the robustness of NER. \vspace{-0.5em}}
	\label{fig:augment} 
\end{figure}

\paragraph{Methods.} The most straightforward method to improve NER robustness is to augment our examples used for training models.
Here we use three intuitive data augmentation methods for the analysis.
\textbf{1)} \texttt{\textbf{Entity Switching}}:
we replace each entity in the target sentence with a different entity of the same type from another sentence.
\textbf{2)} \texttt{\textbf{Random Masking}}:
for each entity, we replace every one of its letters with a random one. We retain the same capitalization pattern and keep all stopwords unchanged. 
\textbf{3)} \texttt{\textbf{Mixing up}}:
inspired by \citet{Guo2019AugmentingDW}, we randomly pick one entity from the target sentence and find another sentence that includes an entity of the same type; then we generate an adversarial sentence by merging the first half of the target sentence (up to and including the entity) with the second half of the second sentence (everything after the entity).
They are illustrated with examples in Figure~\ref{fig:augment}.

\paragraph{Results \& Analysis.}
The results of the three methods on the RoBERTa-CRF model are shown in Table~\ref{tab:results}.
Surprisingly, the most straightforward method, \texttt{Random Masking}, offers the best improvement against entity-level attacks. 
We conjecture it is because it provides more entity patterns, which enhances its entity-level generalization ability and makes models focus more on the context for inference, resulting in a better performance on entity-level attacks (63.4\% $\rightarrow$ 66.3\%). 
As the \texttt{Entity Switching} repeats original entities in the different context of the training set, it aims to improve the performance in using context to infer entities.
The entity-level attacks are indeed better handled (63.4\% $\rightarrow$ 64.7\%).
The \texttt{Mixing up} method, however,  loses the robustness on all settings, possibly due to potential noise from sentences that are not syntactically valid.

\section{Related Work}
\label{app:related}
There are other recent works which also turn their attention from achieving a new state-of-the-art of NER model towards studying NER models' robustness and generalization ability.
\citet{Agarwal2020EntitySwitchedDA} create entity-switched datasets by replacing entities with others of the same type but different national origin.
They find that NER models perform worse on entities from certain countries.
\citet{Mayhew2020RobustNE} and \citet{bodapati-etal-2019-robustness} focus on the robustness when inputs are not written in the standard casing (e.g., ``he is from \textit{us}'' $\rightarrow$ ``US''). 
\citet{Fu2020RethinkingGO} analyze the generalization ability of current NER models by evaluating them across datasets.
\citet{Agarwal2020InterpretabilityAF} further analyze the roles of context and names in entity predictions made by  models and humans.
Although these works begin to understand the robustness issue of NER models, they do not build an automated pipeline to generate natural adversarial instances with large coverage (e.g. thousands of fine-grained classes) at scale.

There are also works in other domains aiming to evaluate models' robustness with perturbed inputs.
For example, \citet{jia2017adversarial} attack reading comprehension models by adding word sequences to the input.
\citet{gan2019improving} and \citet{iyyer-etal-2018-adversarial} paraphrase the input to test models' over-sensitivity.
\citet{jones-etal-2020-robust} target adversarial typos.
\citet{si2020benchmarking} propose a benchmark for reading comprehension with diverse types of test-time perturbation.
These works focus on different domains than our research does, and they do not consider the composition of NER examples.
Little attention is drawn to the entities in the sentences, and many attacks (e.g. character swapping, word injection) may make the perturbed sentences invalid.
To the best of our knowledge, this work is among the first to propose a straightforward, dedicated pipeline for generating natural adversarial examples for the NER task, which takes into account the compositions of NER examples ---\textit{ i.e.,} entity structures and their context.

\section{Conclusion}
\label{sec:conclusion}

Our contributions in this short paper are two-fold.
1) resource-wise: we develop RockNER, a straightforward method for generating natural adversarial attacks for NER, which produces \texttt{\textbf{OntoRock}}, a benchmark for auditing the robustness of NER models. 
2) evaluation-wise: our experimental results and analysis provides answers supported by experimental results to three main research questions on the robustness of current mainstream NER models.
We believe \textit{RockNER} and its produced attacks (e.g., the \texttt{OntoRock} benchmark) can benefit the community working to increase the robustness and out-of-distribution generalization of NER.\footnote{
We leave our full results, more implementation details, and additional analyses in the appendix.}

\section*{Acknowledgements}
This research is supported in part by the Office of the Director of National Intelligence (ODNI), Intelligence Advanced Research Projects Activity (IARPA), via Contract No. 2019-19051600007, the DARPA MCS program under Contract No. N660011924033, the Defense Advanced Research Projects Agency with award W911NF-19-20271, NSF IIS 2048211, NSF SMA 1829268, and gift awards from Google, Amazon, JP Morgan and Sony.
We would like to thank the reviewers for their constructive feedback.




 


\bibliography{citations_rebiber} 
\bibliographystyle{acl_natbib}

\clearpage
\appendix

\section{Statistics of the Entity Dictionary}
\label{app:dictionary_stats}

In Table~\ref{tab:entdictstat}, we show statistics of the adversarial entity dictionary built for the test set.
We generate 279,290 adversarial entities out of 7,433 original entities.
The amount of generated entities is 36 times larger than the original ones, which extremely enriches candidates for conducting entity-level attacks.
Moreover, there is one class for every 2$\sim$10 entities according to their type, and each class includes hundreds of adversarial entities.
This indicates that we have enough adversarial entities to conduct entity-level attacks.

\begin{table}[ht]
	\centering
	\scalebox{0.8
	}{
		\begin{tabular}{@{}c|ccc@{}}
\toprule
\textbf{Type}   & \# \textbf{Original} & \# \textbf{Classes} & \# \textbf{Adversarial} \\ \midrule
GPE           & 2,203           & 237           & 42,912                  \\
ORG           & 1,750           & 402           & 75,259                  \\
FAC           & 135            & 84            & 29,041                  \\
LOC           & 179            & 75            & 20,354                  \\
NORP          & 830            & 117           & 27,033                  \\
LAW           & 40             & 14            & 3,920                   \\
EVENT         & 63             & 33            & 9,636                   \\
WOA & 165            & 73            & 39,508                  \\
PRODUCT       & 74             & 48            & 17,986                  \\
LANG      & 22             & 7             & 4,119                   \\
PERSON        & 1,972           & N/A           & 9,522 \\ \midrule
Total        & 7,433            & 1,090        & 279,290  \
\\\bottomrule         
\end{tabular}
	} 
	\caption{Statistics of the adversarial entity dictionary.}
	\label{tab:entdictstat}
\end{table}

\section{Statistics of the Dataset}
\label{app:dataset_stats}

\begin{table}[h]
	\centering
	\scalebox{0.85
	}{
		\begin{tabular}{c|c|cc}
\toprule
& \textbf{Train}   & \textbf{Dev}    & \textbf{Test}   \\ \midrule
\# Sentences & 59,924   & 8,528   & 8,262   \\
\# Tokens & 1.1M & 148k& 153k \\
\# Entities & 55,008   & 7,482   & 7,433   \\
\# Attacked Entities & N/A & 6,962   & 6,939   \\
\% Attacked Entities & N/A & 93.05  & 93.35 \\
\# Attacked Context Words & N/A &16,155 &15,664   \\
\% Attacked Sentences & N/A &98.03 &97.53 
\\\bottomrule
\end{tabular}
	} 
	\caption{Overall statistics of \texttt{OntoRock} benchmark.}
	\label{tab:overallstat}
\end{table}

\begin{table}[ht]
	\centering
	\scalebox{0.6
	}{
		\begin{tabular}{c|c|cc|cc}
\toprule
 & \textbf{Train}            & \multicolumn{2}{c|}{\textbf{OntoNotes Test}} & \multicolumn{2}{c}{\textbf{\texttt{OntoRock} Test}}                                              \\ \cline{2-6} 
 & \# ent\_words & \# ent\_words  & seen (\%)  & \# ent\_words & seen (\%) \\ \midrule
GPE           & 1615  & 461  & 75.92 & 1202 & 17.30 \\
ORG           & 4037  & 1056 & 67.42 & 2399 & 26.18 \\
FAC           & 681   & 125  & 55.20 & 287  & 19.16 \\
LOC           & 527   & 118  & 66.95 & 224  & 22.32 \\
NORP          & 565   & 160  & 78.13 & 330  & 8.18  \\
LAW           & 343   & 74   & 45.95 & 116  & 26.72 \\
EVENT         & 439   & 91   & 64.84 & 132  & 23.48 \\
WOA & 1107  & 264  & 37.12 & 351  & 17.95 \\
PERSON        & 5367  & 1102 & 61.62 & 1011 & 31.95 \\
PRODUCT       & 381   & 54   & 44.44 & 158  & 7.59  \\
LANGUAGE      & 33    & 7    & 71.43 & 25   & 4.00  \\ \midrule
ALL\_TYPES    & 12174 & 3028 & 68.13 & 5522 & 31.44 \\ \bottomrule
\end{tabular}
	} 
	\caption{Entity statistics of OntoNotes and \texttt{OntoRock} benchmarks.}
	\label{tab:outofdist}
\end{table}

We adopt the train/dev/test splits of OntoNotes used by \citet{pradhan2013towards} in our experiments.
Table \ref{tab:overallstat} presents the statistics of our \texttt{OntoRock} benchmark, which consists of the original OntoNotes training set and our attacked (full version) development and test sets.
Table \ref{tab:outofdist} shows the statistics of entities in the training set, OntoNotes' test set and \texttt{OntoRock}'s test set.
\section{Model Details}
\label{app:model_details}
For spaCy, we load the “en\_core\_web\_lg” model with the white-space tokenizer.

For the Stanza model, we use the English model and set processors as “tokenize, ner” with tokenize\_pretokenized= True.

When we train the Flair model with a GPU, we set mini\_batch\_size as 64, train\_with\_dev as False and embeddings\_storage\_mode as “none”. 

For training BLSTM-CRF, BERT-CRF and RoBERTa-CRF models, we set batch\_size as 20.
We use early stopping and set patience=10 for BLSTM-CRF and 5 for the other two.

\section{Full Results}
\label{app:full_results}

Precision/Recall/F1 scores for each model on the original OntoNotes and our \texttt{OntoRock} benchmark are presented in Table \ref{tab:test_results} (test set) and Table \ref{tab:dev_results} (development set).

\begin{table*}[th!]
	\centering
	\scalebox{0.75
	}{
		\begin{tabular}{c|c|cc||cc||cc||cc}
\toprule
\multicolumn{2}{c|}{\multirow{2}{*}{\begin{tabular}[c]{@{}c@{}}Robustness Analysis \\  (by prediction)\end{tabular}}} &
  \multicolumn{2}{c||}{RoBERTa-CRF  (\%)} &
  \multicolumn{2}{c||}{RB-CRF+\textbf{ES}  (\%)} &
  \multicolumn{2}{c||}{RB-CRF+\textbf{RM} (\%)} &
  \multicolumn{2}{c}{RB-CRF+\textbf{M} (\%)} \\
\multicolumn{2}{c|}{} &
  Attacked &
  Unattacked &
  Attacked &
  Unattacked &
  Attacked &
  Unattacked &
  Attacked &
  Unattacked \\ \midrule
\multirow{4}{*}{\begin{tabular}[c]{@{}c@{}}Correct \\  Type\end{tabular}} &
  d=0 (SameSpan) &
  62.55 &
  86.44 &
  64.78 &
  83.40 &
  66.41 &
  86.23 &
  60.59 &
  84.41 \\ \cline{2-2}
 &
  \begin{tabular}[c]{@{}c@{}}d=1\end{tabular} &
  6.46 &
  0.20 &
  5.76 &
  0.00 &
  6.49 &
  0.20 &
  7.09 &
  1.42 \\ \cline{2-2}
 &
  \begin{tabular}[c]{@{}c@{}}d=2\end{tabular} &
  1.02 &
  0.61 &
  0.60 &
  1.01 &
  1.07 &
  0.81 &
  0.73 &
  1.01 \\ \cline{2-2}
 &
  \begin{tabular}[c]{@{}c@{}}d$\geqslant$3\end{tabular} &
  1.77 &
  1.01 &
  1.69 &
  2.53 &
  1.75 &
  1.21 &
  1.51 &
  1.42 \\ \cline{1-2}
\multirow{4}{*}{\begin{tabular}[c]{@{}c@{}}Wrong \\  Type\end{tabular}} &
 d=0 (SameSpan)&
  14.27 &
  4.45 &
  14.20 &
  4.45 &
  12.88 &
  4.25 &
  14.43 &
  4.25 \\ \cline{2-2}
 &
  \begin{tabular}[c]{@{}c@{}}d=1\end{tabular} &
  1.69 &
  0.00 &
  1.50 &
  0.00 &
  1.59 &
  0.00 &
  1.77 &
  0.00 \\ \cline{2-2}
 &
  \begin{tabular}[c]{@{}c@{}}d=2\end{tabular} &
  0.59 &
  0.20 &
  0.72 &
  0.00 &
  0.55 &
  0.20 &
  0.59 &
  0.20 \\ \cline{2-2}
 &
  \begin{tabular}[c]{@{}c@{}}d$\geqslant$3\end{tabular} &
  0.83 &
  0.00 &
  1.19 &
  0.10 &
  1.15 &
  0.20 &
  1.28 &
  0.20 \\ \cline{1-2}
\multicolumn{2}{c|}{No Prediction} &
  10.82 &
  7.09 &
  9.57 &
  8.50 &
  8.10 &
  6.88 &
  12.02 &
  7.09 \\ \bottomrule
\end{tabular}
	} 
	\caption{Error analysis of RoBERTa-CRF and with three different data augmentation methods. \textit{d} indicates the number of different indices between ground-truth entity and the overlapped predictions (\textbf{ES} for Entity Switching, \textbf{RM} for Random Masking, and \textbf{M} for Mixing up).}
	\label{tab:error_analysis}
\end{table*}

\section{Cases}
\label{app:cases}

\begin{figure*}[ht]
	\centering 
	\includegraphics[width=0.98\linewidth]{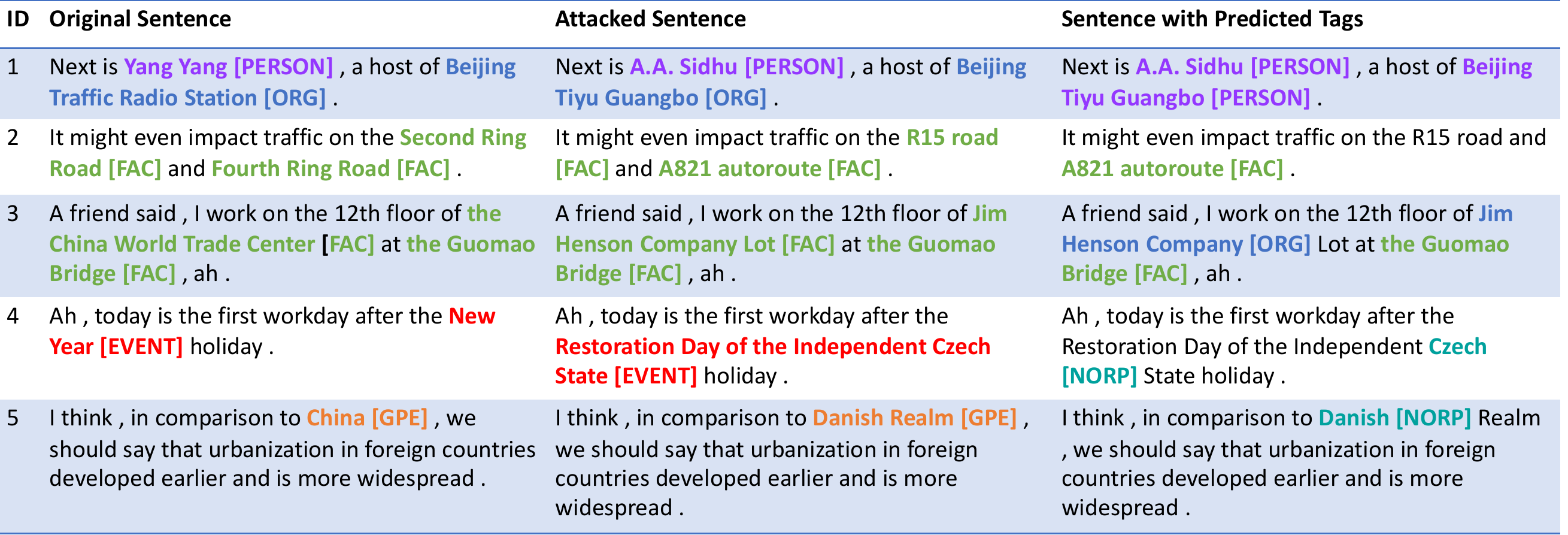}
	\caption{Example cases for the entity-level attacks.}
	\label{fig:cases} 
\end{figure*}

In Fig.~\ref{fig:cases}, we show examples of entity-level attacks on the RoBERTa. 
These examples should be easily solved based on the context.
For example, "a host of" in sentence $1$ and "holiday" in sentence $4$ are both explicit clues.
If NER models are capable of inferring from context, those clues could have assisted them to achieve better performance.
It qualitatively validates our hypothesis that NER models tend to remember entity patterns instead of inferring entity labels from context.

\begin{figure}[ht]
	\centering 
	\includegraphics[width=1\linewidth]{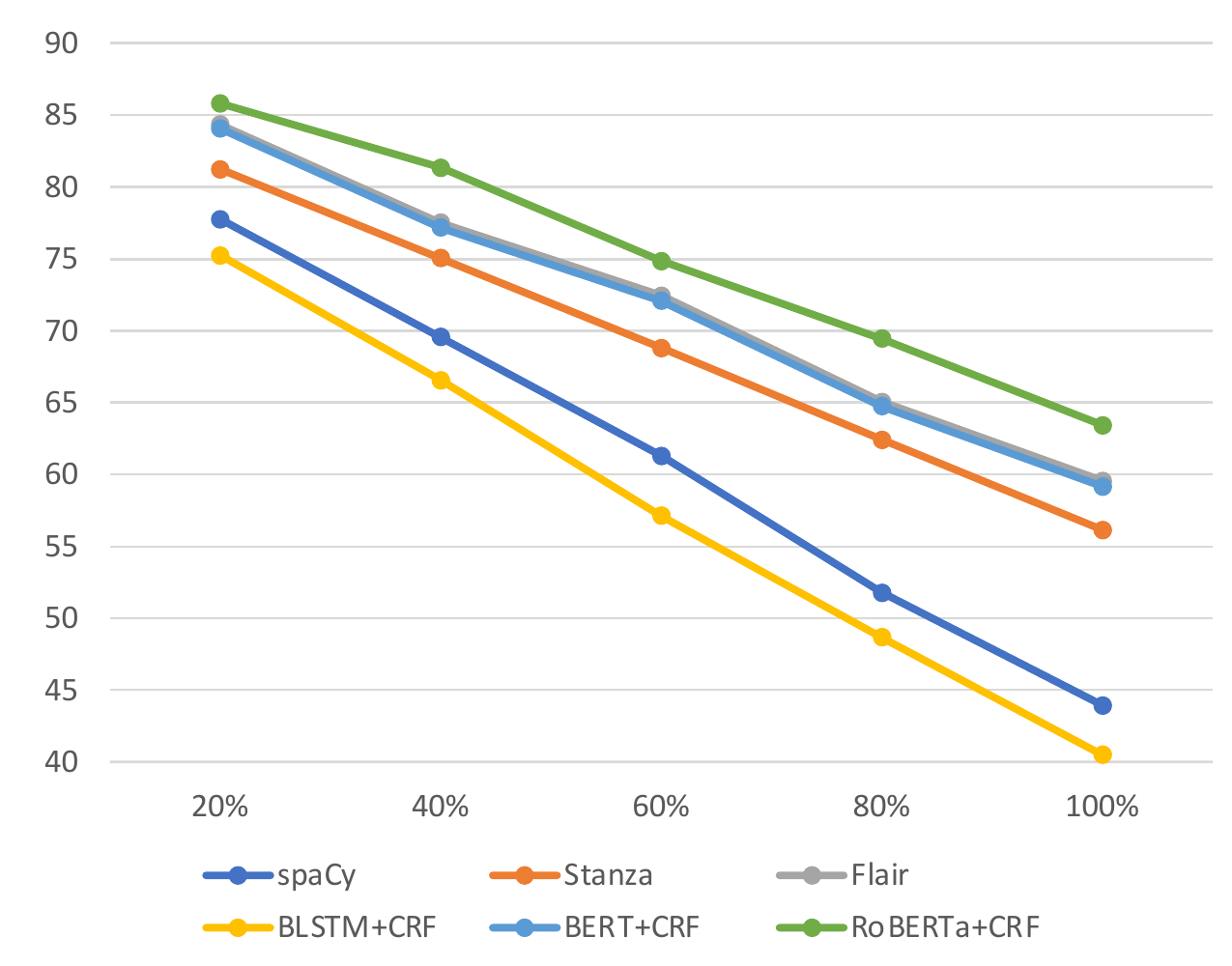}
	\caption{F1 scores for the RoBERTa-CRF model under entity-level attacks with different coverage.}
	\label{fig:curve} 
\end{figure}

\section{More Details of Error Analysis}

\begin{figure*}[th!]
\centering
\begin{subfigure}{.5\textwidth}
  \centering
	\includegraphics[width=0.95\linewidth]{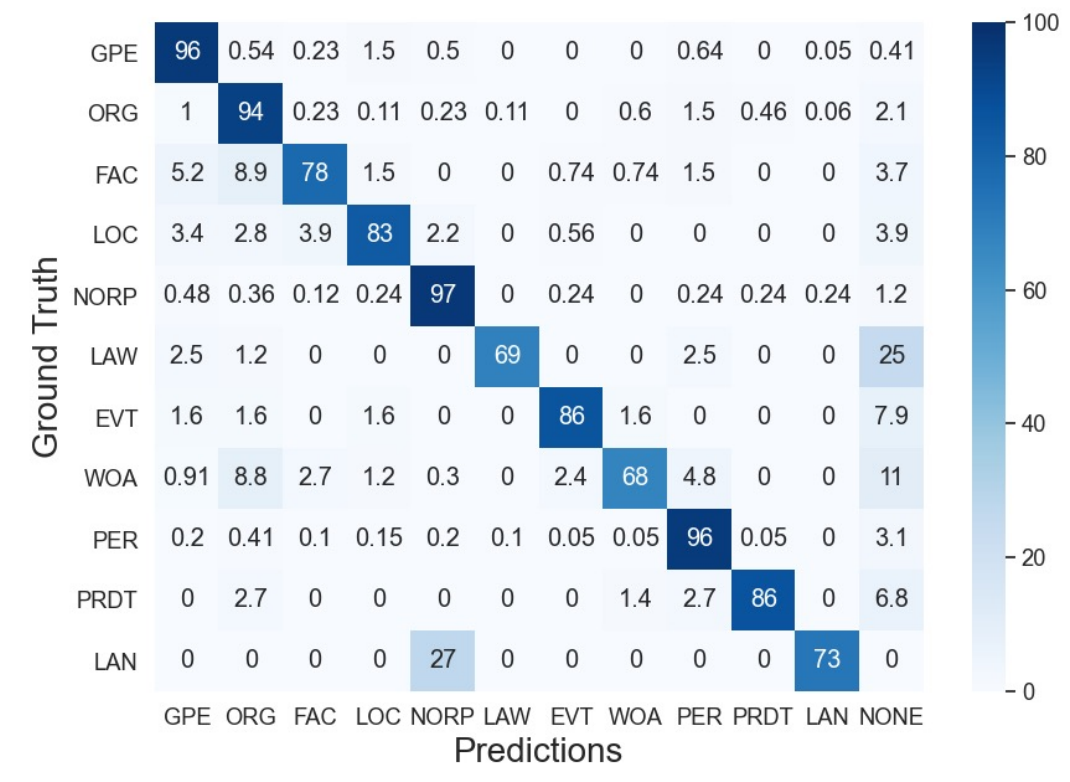}
	\caption{Confusion matrix for ground truth and predictions \\ of RoBERTa-CRF model on the OntoNotes test set. 
    }
	\label{fig:testhm} 
\end{subfigure}%
\begin{subfigure}{.5\textwidth}
  \centering
	\includegraphics[width=0.95\linewidth]{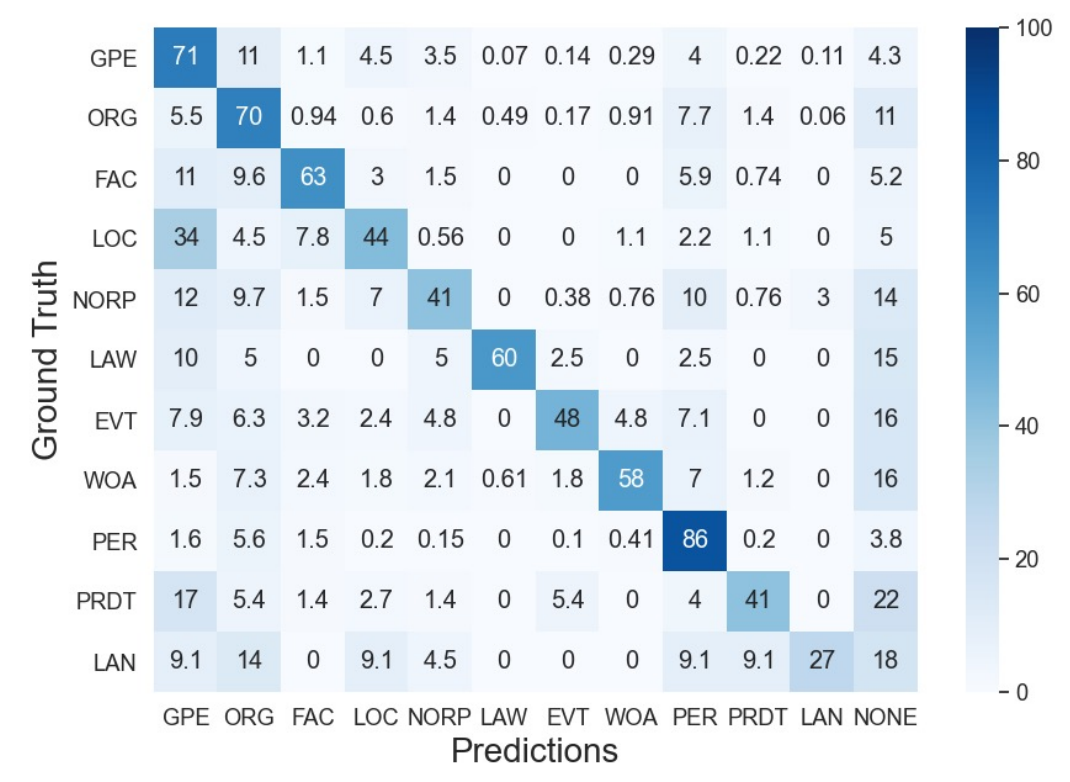}
	\caption{Confusion matrix for ground truth and predictions of RoBERTa-CRF model on the \texttt{OntoRock} test set. 
	}
	\label{fig:rockhm} 
\end{subfigure}
\caption{Confusion matrices for RoBERTa-CRF on OntoNotes' and \texttt{OntoRock}'s test sets.}
\label{fig:confusion}
\end{figure*}

In Figure \ref{fig:confusion}, we present the confusion matrices for RoBERTa-CRF model on the OntoNotes' and \texttt{OntoRock}'s teset sets. We use them to calculate the confusion difference matrix (Figure \ref{fig:error_analysis} (right)).

Following Figure \ref{fig:error_analysis} (left), we categorize the error cases of predictions by the RoBERTa-CRF model and its variants that are trained with augmented data.
The results are presented in Table \ref{tab:error_analysis}.
Among three augmentation methods, random masking gets the highest F1 score on both attacked and original test sets.
The robustness gain mainly comes from more accurate typing (Wrong Type, $d=0$: 14.27\% $\rightarrow$ 12.88\%).

\begin{table*}[t]
	\centering
	\scalebox{0.77
	}{
		\begin{tabular}{@{}r||c|cc|c||c|cc|c||c|cc|c@{}}
\toprule
\textbf{Evaluation Metrics} $\longrightarrow$ & \multicolumn{4}{c||}{Precision (\%)} & \multicolumn{4}{c||}{Recall (\%)} & \multicolumn{4}{c}{F1 (\%)}\\ \midrule
\textbf{Models} $\downarrow$ \textbf{Attack Methods} $\rightarrow$& {\MyColorBox[gray!20]{\textbf{None}}} & {\MyColorBox[red!10]{\textbf{E}}} & {\MyColorBox[cyan!20]{\textbf{C}}} & {\MyColorBox[green!20]{\textbf{E+C}}} & {\MyColorBox[gray!20]{\textbf{None}}} & {\MyColorBox[red!10]{\textbf{E}}} & {\MyColorBox[cyan!20]{\textbf{C}}} & {\MyColorBox[green!20]{\textbf{E+C}}} & {\MyColorBox[gray!20]{\textbf{None}}} & {\MyColorBox[red!10]{\textbf{E}}} & {\MyColorBox[cyan!20]{\textbf{C}}} & {\MyColorBox[green!20]{\textbf{E+C}}}\\ 
\midrule
BLSTM-CRF~\cite{Lample2016NeuralAF} & 86.3 & 40.8 & 78.7 & 32.7 & 82.9 & 40.2 & 76.1 & 32.0 & 84.6 & 40.5 & 77.3 & 32.4\\
\midrule
spaCy~\cite{spacy} &88.3 & 43.9 & 82.0 & 40.1 & 86.2 & 44.0 & 81.6 & 40.1 & 87.3 & 43.9 & 81.8 & 40.1\\ 
Stanza~\cite{qi2020stanza} & 89.5 & 55.3 & 83.9 & 51.2 & 86.3 & 57.1 & 82.1 & 52.2 & 87.9 & 56.1 & 83.0 & 51.7 \\ \midrule
BERT-CRF~\cite{Devlin2019} & 91.9 & 59.8 & 86.7 & 55.7 & 89.3 & 58.5 & 84.9 & 53.5 & 90.6 & 59.2 & 85.8 & 54.6\\ 
Flair~\cite{akbik2018coling} & 91.3 & 59.4 & 86.0 & 55.3 & 90.2 & 59.8 & 86.1 & 55.2 & 90.7 & 59.6 & 86.1 & 55.3\\
\underline{RoBERTa-CRF}~\cite{Liu2019RoBERTaAR} & {\MyColorBox[gray!20]{\textbf{92.9}}} & {\MyColorBox[red!10]{\textbf{63.1}}} & {\MyColorBox[cyan!10]{\textbf{87.4}}} & {\MyColorBox[green!20]{\textbf{58.5}}} & {\MyColorBox[gray!20]{\textbf{91.8}}} & {\MyColorBox[red!10]{\textbf{63.7}}} & {\MyColorBox[cyan!10]{\textbf{87.0}}} & {\MyColorBox[green!20]{\textbf{58.5}}} & {\MyColorBox[gray!20]{\textbf{92.4}}} & {\MyColorBox[red!10]{\textbf{63.4}}} & {\MyColorBox[cyan!10]{\textbf{87.2}}} & {\MyColorBox[green!20]{\textbf{58.5}}}\\ \midrule
\underline{RB-CRF}+\textbf{Entity Switching} & 92.2 & 64.2 & 85.7 & 58.7 & 90.6 & 65.2 & 85.7 & 59.5 & 91.4 & 64.7 & 85.7 & 59.1\\ 
\underline{RB-CRF}+\textbf{Random Masking} & {\MyColorBox[gray!20]{\textbf{92.8}}} & {\MyColorBox[red!10]{\textbf{65.3}}} & 86.1 & {\MyColorBox[green!20]{\textbf{59.2}}} & {\MyColorBox[gray!20]{\textbf{92.4}}} & {\MyColorBox[red!10]{\textbf{67.3}}} & 86.8 & {\MyColorBox[green!20]{\textbf{60.8}}} & {\MyColorBox[gray!20]{\textbf{92.6}}} & {\MyColorBox[red!10]{\textbf{66.3}}} & 86.4 & {\MyColorBox[green!20]{\textbf{60.0}}} \\ 
\underline{RB-CRF}+\textbf{Mixing Up} & 92.5 & 60.7 & {\MyColorBox[cyan!10]{\textbf{86.7}}} & 56.3 & 91.4 & 61.5 & {\MyColorBox[cyan!10]{\textbf{87.0}}} & 56.8 & 92.0 & 61.1 & {\MyColorBox[cyan!10]{\textbf{86.9}}} & 56.5\\ \bottomrule             
\end{tabular}
	} 
	\caption{Results of NER models on the \textbf{test} set of OntoNotes with {\MyColorBox[gray!20]{\textbf{none}}} changes and three variants of the \texttt{OntoRock} benchmark ({\MyColorBox[red!10]{\textbf{E}}} for entity-only attacks, {\MyColorBox[cyan!20]{\textbf{C}}} for context-only attacks, and {\MyColorBox[green!20]{\textbf{E+C}}} for the full version).}
	\label{tab:test_results}
\end{table*}

\begin{table*}[t]
	\centering
	\scalebox{0.77
	}{
		\begin{tabular}{@{}r||c|cc|c||c|cc|c||c|cc|c@{}}
\toprule
Evaluation Metrics $\longrightarrow$ & \multicolumn{4}{c||}{Precision (\%)} & \multicolumn{4}{c||}{Recall (\%)} & \multicolumn{4}{c}{F1 (\%)}\\ \midrule
Models $\downarrow$ Attack Methods $\rightarrow$& {\MyColorBox[gray!20]{\textbf{None}}} & {\MyColorBox[blue!10]{\textbf{E}}} & {\MyColorBox[cyan!20]{\textbf{C}}} & {\MyColorBox[green!20]{\textbf{E+C}}} & {\MyColorBox[gray!20]{\textbf{None}}} & {\MyColorBox[blue!10]{\textbf{E}}} & {\MyColorBox[cyan!20]{\textbf{C}}} & {\MyColorBox[green!20]{\textbf{E+C}}} & {\MyColorBox[gray!20]{\textbf{None}}} & {\MyColorBox[blue!10]{\textbf{E}}} & {\MyColorBox[cyan!20]{\textbf{C}}} & {\MyColorBox[green!20]{\textbf{E+C}}}\\ 
\midrule
BLSTM-CRF~\cite{Lample2016NeuralAF} & 85.4 & 40.0 & 77.2 & 33.2 & 82.5 & 39.9 & 75.2 & 32.6 & 83.9 & 40.0 & 76.2 & 32.9\\ 
\midrule
spaCy~\cite{spacy} & 86.0 & 43.2 & 79.7 & 40.3 & 84.8 & 44.0 & 79.7 & 40.4 & 85.4 & 43.6 & 79.7 & 40.3\\ 
Stanza~\cite{qi2020stanza} & 87.5 & 52.6 & 81.1 & 48.8 & 84.2 & 55.2 & 79.2 & 50.8 & 85.8 & 53.9 & 80.1 & 49.8 \\ \midrule
BERT-CRF~\cite{Devlin2019} & 91.2 & 58.1 & 84.8 & 54.3 & 88.9 & 57.3 & 83.1 & 52.7 & 90.0 & 57.7 & 84.0 & 53.5\\ 
Flair~\cite{akbik2018coling} & 89.4 & 56.6 & 83.6 & 52.5 & 89.0 & 58.1 & 84.1 & 53.2 & 89.2 & 57.3 & 83.9 & 52.9\\
\underline{RoBERTa-CRF}~\cite{Liu2019RoBERTaAR} & {\MyColorBox[gray!20]{\textbf{91.3}}} & {\MyColorBox[blue!10]{\textbf{60.9}}} & {\MyColorBox[cyan!10]{\textbf{84.9}}} & {\MyColorBox[green!20]{\textbf{56.1}}} & {\MyColorBox[gray!20]{\textbf{90.4}}} & {\MyColorBox[blue!10]{\textbf{62.2}}} & {\MyColorBox[cyan!10]{\textbf{84.8}}} & {\MyColorBox[green!20]{\textbf{56.9}}} & {\MyColorBox[gray!20]{\textbf{90.0}}} & {\MyColorBox[blue!10]{\textbf{61.6}}} & {\MyColorBox[cyan!10]{\textbf{84.8}}} & {\MyColorBox[green!20]{\textbf{56.5}}}\\ \midrule
\underline{RB-CRF}+\textbf{Entity Switching} & 90.3 & 61.8 & 83.5 & 56.1 & 89.5 & 64.5 & 83.8 & 58.1 & 89.9 & 63.1 & 83.7 & 57.1\\ 
\underline{RB-CRF}+\textbf{Random Masking} & 90.6 & {\MyColorBox[blue!10]{\textbf{62.5}}} & 83.7 & {\MyColorBox[green!20]{\textbf{56.4}}} & {\MyColorBox[gray!20]{\textbf{90.7}}} & {\MyColorBox[blue!10]{\textbf{65.2}}} & 84.9 & {\MyColorBox[green!20]{\textbf{58.5}}} & {\MyColorBox[gray!20]{\textbf{90.7}}} & {\MyColorBox[blue!10]{\textbf{63.8}}} & 84.3 & {\MyColorBox[green!20]{\textbf{57.4}}} \\ 
\underline{RB-CRF}+\textbf{Mixing Up} & {\MyColorBox[gray!20]{\textbf{90.8}}} & 58.1 & {\MyColorBox[cyan!10]{\textbf{84.4}}} & 53.5 & 90.6 & 60.0 & {\MyColorBox[cyan!10]{\textbf{85.5}}} & 55.2 & {\MyColorBox[gray!20]{\textbf{90.7}}} & 59.0 & {\MyColorBox[cyan!10]{\textbf{85.0}}} & 54.4\\ \bottomrule             
\end{tabular}
	} 
	\caption{Results of NER models on the \textbf{development} set of OntoNotes with {\MyColorBox[gray!20]{\textbf{none}}} changes and three variants of the \texttt{OntoRock} benchmark ({\MyColorBox[red!10]{\textbf{E}}} for entity-only attacks, {\MyColorBox[cyan!20]{\textbf{C}}} for context-only attacks, and {\MyColorBox[green!20]{\textbf{E+C}}} for the full version).}
	\label{tab:dev_results}
\end{table*}

\section{Attacking Curve}

For the entity-level attacks, we conduct 5 separate attacks by replacing 20\%, 40\%, 60\%, 80\%, and 100\% of the entities in the test set.
For each model, we evaluate it on the 5 generated test sets and plot a curve of the F1 scores for each attack, shown in Fig.~\ref{fig:curve}.
The descending trend is intuitive, and the performance of weak models drops more rapidly than the performance of strong models.

\end{document}